\documentclass[sigconf]{acmart}

\usepackage{multirow}
\usepackage{tabulary}

\usepackage{makecell}

\AtBeginDocument{%
  \providecommand\BibTeX{{%
    \normalfont B\kern-0.5em{\scshape i\kern-0.25em b}\kern-0.8em\TeX}}}

\copyrightyear{2021}
\acmYear{2021}
\setcopyright{acmcopyright}\acmConference[MM '21]{Proceedings of the 29th ACM International Conference on Multimedia}{October 20--24, 2021}{Virtual Event, China}
\acmBooktitle{Proceedings of the 29th ACM International Conference on Multimedia (MM '21), October 20--24, 2021, Virtual Event, China}
\acmPrice{15.00}
\acmDOI{10.1145/3474085.3478331}
\acmISBN{978-1-4503-8651-7/21/10}

\begin{document}
\fancyhead{}

\title{\texttt{X-modaler}: A Versatile and High-performance Codebase\\ for Cross-modal Analytics}

\author{Yehao Li, Yingwei Pan, Jingwen Chen, Ting Yao, and Tao Mei}
\affiliation{%
  \institution{JD AI Research, Beijing, China}}
\email{{yehaoli.sysu,panyw.ustc,chenjingwen.sysu,tingyao.ustc}@gmail.com;tmei@jd.com}

%
\renewcommand{\shortauthors}{Li, et al.}

%
\begin{abstract}

With the rise and development of deep learning over the past decade, there has been a steady momentum of innovation and breakthroughs that convincingly push the state-of-the-art of cross-modal analytics between vision and language in multimedia field. Nevertheless, there has not been an open-source codebase in support of training and deploying numerous neural network models for cross-modal analytics in a unified and modular fashion. In this work, we propose \texttt{X-modaler} --- a versatile and high-performance codebase that encapsulates the state-of-the-art cross-modal analytics into several general-purpose stages (e.g., pre-processing, encoder, cross-modal interaction, decoder, and decode strategy). Each stage is empowered with the functionality that covers a series of modules widely adopted in state-of-the-arts and allows seamless switching in between. This way naturally enables a flexible implementation of state-of-the-art algorithms for image captioning, video captioning, and vision-language pre-training, aiming to facilitate the rapid development of research community. Meanwhile, since the effective modular designs in several stages (e.g., cross-modal interaction) are shared across different vision-language tasks, \texttt{X-modaler} can be simply extended to power startup prototypes for other tasks in cross-modal analytics, including visual question answering, visual commonsense reasoning, and cross-modal retrieval.
\texttt{X-modaler} is an Apache-licensed codebase, and its source codes, sample projects and pre-trained models are available on-line: https://github.com/YehLi/xmodaler.
\end{abstract}

%
%



%

%
\maketitle

\section{Introduction}

Vision and language are two fundamental capabilities of human intelligence. Humans routinely perform cross-modal analytics through the interactions between vision and language, supporting the uniquely human capacity, such as describing what they see with a natural sentence (image captioning \cite{anderson2017bottom,yao2018exploring,yao2019hierarchy} and video captioning \cite{pan2016jointly,li2018jointly,Yao:ICCV15}) and answering open-ended questions w.r.t the given image (visual question answering \cite{antol2015vqa,kim2018bilinear}). The valid question of how language interacts with vision motivates researchers to
expand the horizons of multimedia area by exploiting cross-modal analytics in different scenarios. In the past five years, vision-to-language has been one of the ``hottest'' and fast-developing topics for cross-modal analytics, with a significant growth in both volume of publications and extensive applications, e.g., image/video captioning and the emerging research task of vision-language pre-training. Although numerous existing vision-to-language works have released the open-source implementations, the source codes are implemented in different deep learning platforms (e.g., Caffe, TensorFlow, and PyTorch) and most of them are not organized in a standardized and user-friendly manner. Thus, researchers and engineers have to make intensive efforts to deploy their own ideas/applications for vision-to-language based on existing open-source implementations, which severely hinders the rapid development of cross-modal analytics.

To alleviate this issue, we propose the \texttt{X-modaler} codebase, a versatile, user-friendly and high-performance PyTorch-based library that enables a flexible implementation of state-of-the-art vision-to-language techniques by organizing all components in a modular fashion. To our best knowledge, \texttt{X-modaler} is the first open-source codebase for cross-modal analytics that accommodates numerous kinds of vision-to-language tasks.

\begin{figure*}
    \centering
    \vspace{-0.12in}
    \includegraphics[width=0.88\linewidth]{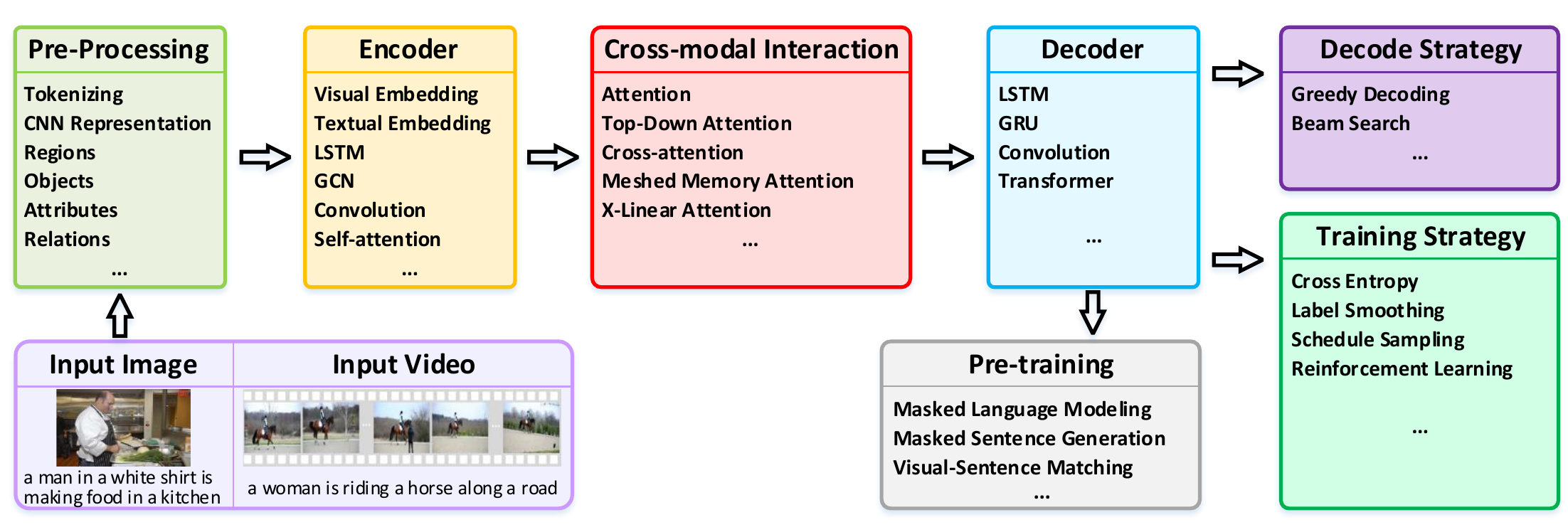}
    \vspace{-0.2in}
    \caption{An overview of the architecture in \texttt{X-modaler}, which is composed of seven major stages: \texttt{pre-processing}, \texttt{encoder}, \texttt{cross-modal interaction}, \texttt{decoder}, \texttt{decode strategy}, \texttt{training strategy}, and \texttt{pre-training}. Each stage is empowered with the functionality that covers a series of widely adopted modules in state-of-the-arts.}
    \vspace{-0.12in}
    \label{fig:architecture}
\end{figure*}

Specifically, taking the inspiration from neural machine translation in NLP field, the typical architecture of vision-to-language models is essentially an encoder-decoder structure. An image/video is first represented as a set of visual tokens (regions or frames), CNN representation, or high-level attributes via \texttt{pre-processing}, which are further transformed into intermediate states via \texttt{encoder} (e.g., LSTM, Convolution, or Transformer-based encoder). Next, conditioned the intermediate states, the \texttt{decoder} is utilized to decode each word at each time step, followed by \texttt{decode strategy} module (e.g., greedy decoding or beam search) to compose the final output sentence. More importantly, recent progress on cross-modal analytics has featured visual attention mechanisms, that trigger the cross-modal interaction between the visual content (transformed by encoder) and the textual sentence (generated by decoder) to boost vision-to-language. Therefore, an additional stage of \texttt{cross-modal interaction} is commonly adopted in state-of-the-art vision-to-language techniques. The whole encoder-decoder structure can be optimized with different \texttt{training strategies} (e.g., cross entropy loss, or reinforcement learning). In addition, vision-language pre-training approaches (e.g., \cite{zhou2019unified,li2021scheduled,pan2020auto}) go beyond the typical encoder-decoder structure by including additional \texttt{pre-training} objectives (e.g., masked language modeling and masked sentence generation). In this way, the state-of-the-art cross-modal analytics techniques can be encapsulated into seven general-purpose stages: \texttt{pre-processing}, \texttt{encoder}, \texttt{cross-modal interaction}, \texttt{decoder}, \texttt{decode strategy}, \texttt{training strategy}, and \texttt{pre-training}. Following this philosophy, \texttt{X-modaler} is composed of these seven general-purpose stages, and each stage is empowered with the functionality that covers a series of commonly adopted modules in state-of-the-arts. Such modular design in \texttt{X-modaler} enables a flexible implementation of state-of-the-art algorithms for vision-to-language, and meanwhile allows the easy plug-ins of user-defined modules to facilitate the deployment of novel ideas/applications for cross-modal analytics.

In summary, we have made the following contributions: \textbf{(I).} To our best knowledge, \texttt{X-modaler} is the first open-source codebase that unifies comprehensive high-quality modules for cross-modal analytics. \textbf{(II).} \texttt{X-modaler} provides the easy implementations of state-of-the-art models for image captioning, video captioning, and vision-language pre-training, in a standardized and user-friendly manner. Moreover, \texttt{X-modaler} can be simply extended to support other vision-language tasks, e.g., visual question answering, visual commonsense reasoning, and cross-modal retrieval. \textbf{(III).} In \texttt{X-modaler}, we release all the reference codes, pre-trained models, and tools for each vision-language task, which will offer a fertile ground for deploying cross-modal analytics in industry and designing novel architectures in academia.

\begin{figure}
    \centering
    \vspace{-0.16in}
    \includegraphics[width=0.86\linewidth]{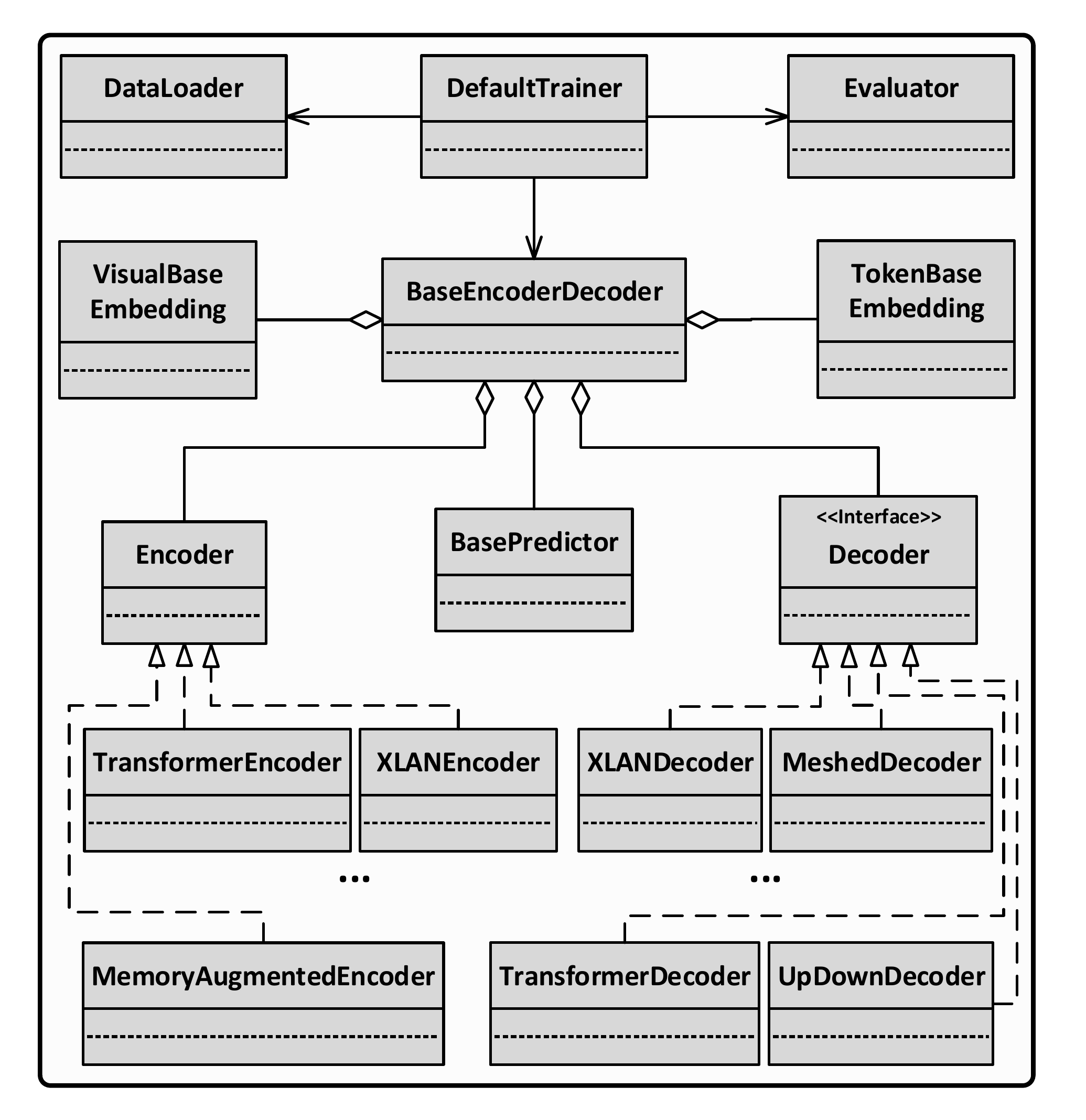}
    \vspace{-0.22in}
    \caption{Class diagram of our \texttt{X-modaler} codebase.}
    \vspace{-0.22in}
    \label{fig:uml}
\end{figure}

\vspace{-0.1in}
\section{Architecture}

In this section, we present the detailed architecture in our \texttt{X-modaler} consisting of seven major stages, as shown in Figure \ref{fig:architecture}. Figure \ref{fig:uml} depicts the detailed class diagram of our \texttt{X-modaler} codebase.

\vspace{-0.1in}

\subsection{Pre-processing}
The \texttt{pre-processing} stage is utilized to transform the primary inputs of image/video and textual sentence into visual and textual tokens. Besides the typical \textbf{tokenizing}, we include numerous modules to represent each input image/video in different ways: (1) directly taking the output \textbf{CNN Representation} of fully-connected layers in 2D/3D CNNs as image/video features; (2) detecting a set of \textbf{regions} as bottom-up signals via Faster R-CNN as in \cite{anderson2017bottom}; (3) recognizing \textbf{objects} from visual content via pre-trained object classifier \cite{yao2017novel}; (4) extracting high-level semantic \textbf{attributes} through Multiple Instance Learning \cite{yao2017boosting,pan2017video}; (5) exploring the semantic or spatial object \textbf{relations} between every two regions \cite{yao2018exploring}.

\vspace{-0.06in}
\subsection{Encoder}
The \texttt{encoder} stage is to take the visual/textual tokens as input and produce intermediate states to encode the semantic content. After transforming each visual/textual token via \textbf{visual/textual embedding}, the most typical way to construct encoder is to directly adopt \textbf{LSTM} to sequentially encode the sequence of tokens. The module of Graph Convolutional Networks (\textbf{GCN}) \cite{yao2018exploring} can be further utilized to strengthen each region-level encoded features by exploiting the graph structure among regions. Instead of the sequential modeling in LSTM module, \textbf{Convolution} module \cite{aneja2018convolutional,chen2019temporal} fully employs convolutions in encoder to enable parallelization within a sequence during training. Inspired by recent success of Transform-style encoder-decoder in NLP field \cite{vaswani2017attention}, we include the \textbf{self-attention} module that leverages self-attention mechanism to enhance each local (region/frame) feature by exploring the intra-modal feature interactions.

\vspace{-0.06in}

\subsection{Cross-modal Interaction}
The \texttt{cross-modal interaction} stage aims to boost vision-language tasks by encouraging more interactions between the two different modalities. \textbf{Attention} module denotes the conventional attention mechanism that dynamically measures the contribution of each local image region \cite{Xu:ICML15} or frame \cite{Yao:ICCV15} based on the hidden state of decoder. \textbf{Top-down attention} module \cite{anderson2017bottom} exploits visual attention at object level. \textbf{Co-attention} module \cite{lu2019vilbert} enables bi-directional interaction between visual and textual tokens. \textbf{Meshed memory attention} module \cite{cornia2020meshed} utilizes memory-augmented attention that boosts inter-modal interaction with a priori knowledge. \textbf{X-Linear attention} module \cite{pan2020x} models the higher order interactions with both spatial and channel-wise bilinear attention.

\vspace{-0.06in}

\subsection{Decoder}
The \texttt{decoder} stage targets for decoding each word at each time step conditioned on the intermediate states of the inputs induced by the encoder. Similar to the modules of encoder stage, the decoder can also be constructed in different forms: \textbf{LSTM/GRU} that autoregressively produces each word, \textbf{Convolution} which fully leverages convolutions for decoding sentence, or \textbf{Transformer} that first exploits the word dependency via self-attention mechanism and further captures the co-attention across vision \& language via cross-attention mechanism to facilitate word generation.

\vspace{-0.06in}

\subsection{Decode Strategy}
The \texttt{decode strategy} stage includes two common modules to generate the final output sentence at inference: (1) \textbf{greedy decoding} that samples the word with the maximum probability at each time step until we select the special end-of-sentence token or reach the maximum length; (2) \textbf{beam search}, i.e., a heuristic search algorithm that maintains a beam including several most likely partial sentences at each decoding time step.

\vspace{-0.06in}
\subsection{Training Strategy}
In the \texttt{training strategy} stage, we assemble a series of practical training strategies adopted in state-of-the-art vision-to-language models: (1) \textbf{cross entropy} module that penalizes the prediction of each word with cross entropy loss; (2) \textbf{label smoothing} module \cite{szegedy2016rethinking} further regularizes the classifier for word prediction by estimating the marginalized effect of label-dropout; (3) \textbf{schedule sampling} module \cite{bengio2015scheduled} capitalizes on a curriculum learning strategy to gently change the input token at each time step from the ground-truth word token to the estimated one from previous step; (4) \textbf{reinforcement learning} module \cite{rennie2017self} enables the direct optimization of the whole encoder-decoder structure with expected sentence-level reward loss.

\vspace{-0.06in}
\subsection{Pre-training}
To endow the base encoder-decoder structure with the capabilities of multi-modal reasoning for vision-language pre-training, we involve the \texttt{pre-training} stage that pre-trains the base structure with several vision-language proxy tasks, e.g., \textbf{masked language modeling}, \textbf{masked sentence generation}, and \textbf{visual-sentence matching} as in \cite{zhou2019unified,li2021scheduled}.

\begin{figure}
    \centering
    \vspace{-0.16in}
    \includegraphics[width=0.86\linewidth]{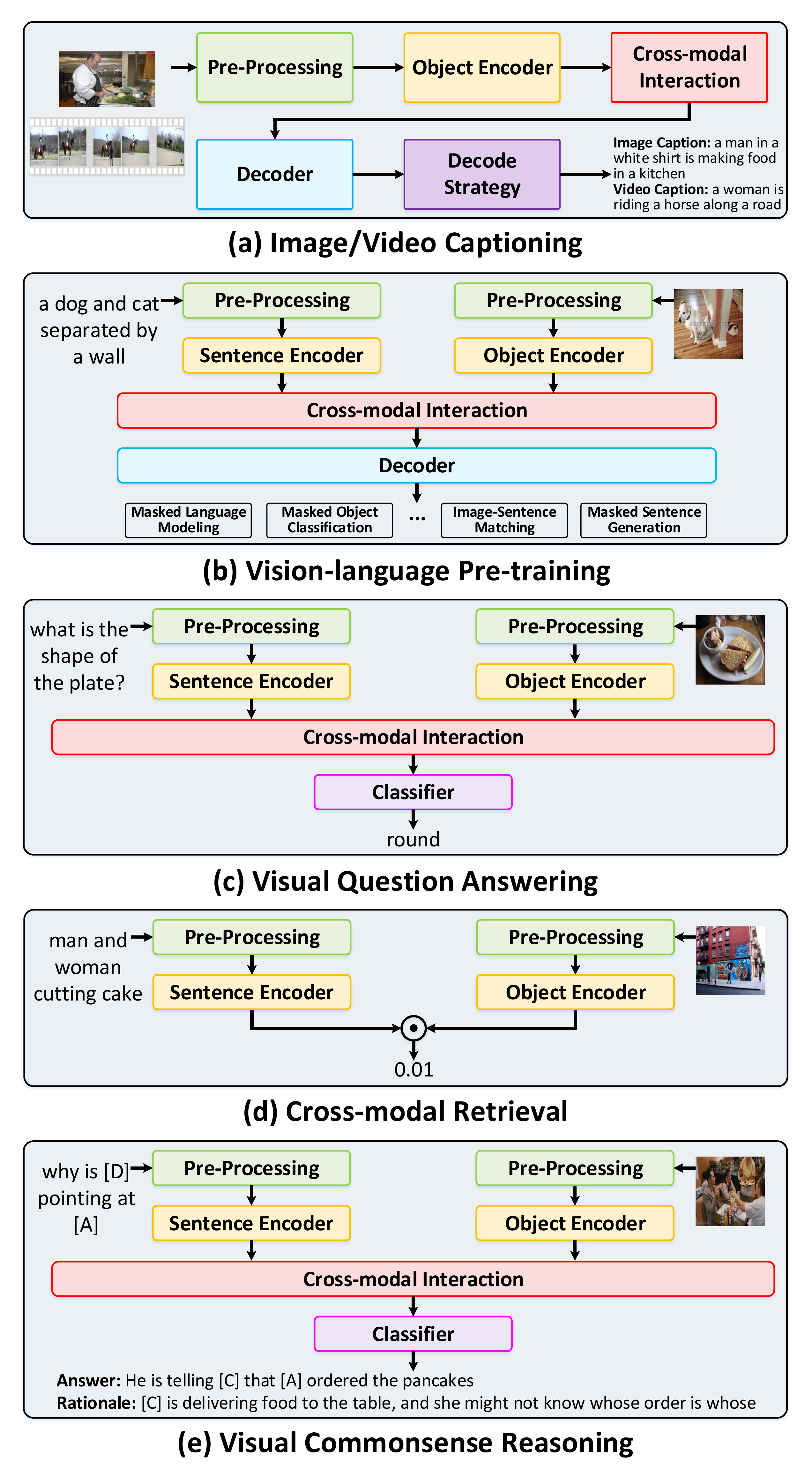}
    \vspace{-0.2in}
    \caption{Exemplary implementations of cross-modal analytics for numerous vision-language tasks in our \texttt{X-modaler}.}
    \vspace{-0.2in}
    \label{fig:task}
\end{figure}

\section{Applications and Evaluations}

This section details the exemplary implementations of each vision-language task (as shown in Figure \ref{fig:task}) in our \texttt{X-modaler}, coupled with the experimental results over several vision-language benchmarks (e.g., COCO \cite{chen2015microsoft}, MSVD \cite{Chen:ACL11}, and MSR-VTT \cite{Xu:CVPR16}) for image/video captioning task.

\noindent\textbf{Image/Video Captioning.}
This task aims to auto-regressively generate the natural sentence that depicts the visual content of input image/video. We re-implement several state-of-the-art image captioning approaches (e.g., $M^2$ Transformer \cite{cornia2020meshed} and X-LAN \cite{pan2020x}) and video captioning methods (e.g., TDConvED \cite{chen2019temporal} and Transformer \cite{sharma2018conceptual}) by allocating different modules in the unified encoder-decoder paradigm (see Figure \ref{fig:task} (a)). Table \ref{tab:COCO} and \ref{tab:video} show the performance comparison of our re-implemented methods through \texttt{X-modaler} over COCO for image captioning and MSVD \& MSR-VTT for video captioning, which manage to achieve state-of-the-art performances on each benchmark.

\begin{table}[t]\small
    \centering
    \vspace{-0.13in}
    \setlength\tabcolsep{1.2pt}
    \caption{\small Performance comparison on COCO for image captioning.}
    \vspace{-0.16in}
    \begin{tabular}{l | c c c c c | c c c c c}
    \Xhline{2\arrayrulewidth}
		  & \multicolumn{5}{c|}{\textbf{Cross-Entropy Loss}} & \multicolumn{5}{c}{\textbf{CIDEr Score Optimization}} \\
		  & B@4  & M    & R    & C     & S    & B@4  & M    & R    & C     & S  \\	\hline \hline
Attention \cite{rennie2017self}            & 36.1 & 27.6 & 56.6 & 113.0 & 20.4 & 37.1 & 27.9 & 57.6 & 123.1 & 21.3 \\
Up-Down \cite{anderson2017bottom}          & 36.0 & 27.6 & 56.6 & 113.1 & 20.7 & 37.7 & 28.0 & 58.0 & 124.7 & 21.5 \\
Transformer \cite{sharma2018conceptual}    & 35.8 & 28.2 & 56.7 & 116.6 & 21.3 & 39.2 & 29.1 & 58.7 & 130.0 & 23.0 \\
$M^2$ Transformer \cite{cornia2020meshed}  & 35.7 & 27.8 & 56.3 & 114.5 & 20.7 & 38.8 & 28.8 & 58.0 & 130.0 & 22.3 \\
X-LAN \cite{pan2020x}                      & 37.5 & 28.6 & 57.6 & 120.7 & 21.9 & 39.2 & 29.4 & 59.0 & 131.0 & 23.2 \\

	\Xhline{2\arrayrulewidth}
    \end{tabular}
	\vspace{-0.22in}
    \label{tab:COCO}
\end{table}

\noindent\textbf{Vision-language Pre-training (VLP).}
VLP is to pre-train a unified encoder-decoder structure over large-scale vision-language benchmarks, which can be easily adapted to vision-language downstream tasks. The state-of-the-art VLP models (e.g., TDEN \cite{li2021scheduled} and ViLBERT \cite{lu2019vilbert}) can be implemented as a two-stream Transformer structure (see Figure \ref{fig:task} (b)): object and sentence encoders first separately learns the representations of each modality, and the cross-modal interaction module further performs multi-modal reasoning, followed with the decoder for sentence generation.

\noindent\textbf{Visual Question Answering (VQA).}
In VQA, the model predicts an answer to the given natural language question with regard to an image. Here we show the implementation of a base model for this task in Figure \ref{fig:task} (c). This base model first encodes the input question and image via object and sentence encoders, and further utilizes cross-modal interaction module (e.g., the attention mechanism in \citep{yu2019deep}) to achieve the holistic image-question representation. Finally, a single-layer MLP is leveraged as a classifier to predict answer based on the holistic image-question representation.

\noindent\textbf{Cross-modal Retrieval.}
It aims to search an image/caption from a pool given its caption/image. It is natural to formulate this task as a ranking problem that sorts images/captions according to the learnt image-sentence matching scores. Here the image-sentence matching score can be directly measured as the dot product between the encoded features of image and caption (see Figure \ref{fig:task} (d)).

\noindent\textbf{Visual Commonsense Reasoning (VCR).}
VCR tackles two problems: visual question answering and answer justification, that requires the model to predict an answer or judge the correctness of the chosen rationale respectively. Each problem is framed as multiple choice task. Similar to VQA, we can measure the holistic image-sentence feature via cross-modal interaction module based on the multi-modal outputs of object and sentence encoders, which will be further leveraged to predict the score for each possible response via a classifier (see Figure \ref{fig:task} (e)).

\section{Conclusions}

We presented \texttt{X-modaler}, a versatile and high-performance codebase for cross-modal analytics. This codebase unifies comprehensive high-quality modules in state-of-the-art vision-language techniques, which are organized in a standardized and user-friendly fashion. Through an extensive set of experiments on several vision-language benchmarks (e.g., COCO, MSVD, and MSR-VTT), we demonstrate that our \texttt{X-modaler} provides state-of-the-art solutions for image/video captioning task. For the ease of use, all the reference codes, pre-trained models, and tools for each vision-language task are published on GitHub.

\begin{table}[t]\small
\centering
\vspace{-0.13in}
    \setlength\tabcolsep{4.1pt}
    \caption{\small Performance comparison for video captioning.}
    \vspace{-0.16in}
\begin{tabular}{l|cccc|cccc}
\Xhline{2\arrayrulewidth}
\multicolumn{1}{c|}{\multirow{2}{*}{Model}} & \multicolumn{4}{c|}{MSVD} & \multicolumn{4}{c}{MSR-VTT} \\
\multicolumn{1}{c|}{}                       & B@4   & M     & R  & C    & B@4   & M     & R    & C    \\ \hline \hline
MP-LSTM \cite{Venugopalan14}                & 48.1  & 32.4  & 68.1 & 73.1 & 38.6 & 26.0  & 58.3 & 41.1 \\
TA  \cite{Yao:ICCV15}                       & 51.0  & 33.5  & 70.0 & 77.2 & 39.9 & 26.4  & 59.4 & 42.9 \\
Transformer \cite{sharma2018conceptual}     & 49.4  & 33.3  & 68.7 & 80.3 & 39.2 & 26.5  & 58.7 & 44.0 \\
TDConvED \cite{chen2019temporal}            & 51.7  & 34.1  & 70.4 & 77.8 & 38.9 & 26.3  & 59.0 & 40.7 \\
\Xhline{2\arrayrulewidth}
\end{tabular}
\vspace{-0.26in}
\label{tab:video}
\end{table}

\end{document}